\documentclass[journal]{IEEEtran}
\usepackage{amsmath,amsfonts}
\usepackage{algorithmic}
\usepackage{algorithm}
\usepackage{array}
\usepackage{makecell}
\usepackage[caption=false,font=normalsize,labelfont=sf,textfont=sf]{subfig}
\usepackage{textcomp}
\usepackage{stfloats}
\usepackage{booktabs}
\usepackage{url}
\usepackage{times}
\usepackage{verbatim}
\usepackage{graphicx}
\usepackage{cite}
\usepackage{multirow}
\usepackage{amsmath}
\begin{document}
	
	\title{Erasure-based Interaction Network for RGBT Video Object Detection and A Unified Benchmark}
	
	\author{Zhengzheng Tu, Qishun Wang, Hongshun Wang, Kunpeng Wang, Chenglong Li~\IEEEmembership{}
		\thanks{This paper was produced by the IEEE Publication Technology Group. They are in Piscataway, NJ.}
		\thanks{Manuscript received April 19, 2021; revised August 16, 2021.}}
	
	\markboth{Journal of \LaTeX\ Class Files,~Vol.~14, No.~8, August~2021}%
	{Shell \MakeLowercase{\textit{et al.}}: A Sample Article Using IEEEtran.cls for IEEE Journals}

	\maketitle

	\begin{abstract}
		Recently, many breakthroughs are made in the field of Video Object Detection (VOD), but the performance is still limited due to the imaging limitations of RGB sensors in adverse illumination conditions. To alleviate this issue, this work introduces a new computer vision task called RGB-thermal (RGBT) VOD by introducing the thermal modality that is insensitive to adverse illumination conditions. To promote the research and development of RGBT VOD, we design a novel Erasure-based Interaction Network (EINet) and establish a comprehensive benchmark dataset (VT-VOD50) for this task. Traditional VOD methods often leverage temporal information by using many auxiliary frames, and thus have large computational burden. Considering that thermal images exhibit less noise than RGB ones, we develop a negative activation function that is used to erase the noise of RGB features with the help of thermal image features. Furthermore, with the benefits from thermal images, we rely only on a small temporal window to model the spatio-temporal information to greatly improve efficiency while maintaining detection accuracy.
		VT-VOD50 dataset consists of 50 pairs of challenging RGBT video sequences with complex backgrounds, various objects and different illuminations, which are collected in real traffic scenarios. Extensive experiments on VT-VOD50 dataset demonstrate the effectiveness and efficiency of our proposed method against existing mainstream VOD methods. The code of EINet and the dataset will be released to the public for free academic usage.
	\end{abstract}
	
	
	\section{Introduction}
	\IEEEPARstart In recent years, Video Object Detection (VOD) \cite{10030892,chen2020memory,he2021end} task has gradually attracted more and more attention. The goal of VOD is to capture the category and location of every object in each frame of the video. It plays a vital role in various applications in traffic scenario, including autonomous driving and road monitoring. However, the performances of existing VOD methods are limited by single RGB imaging and cannot perform well under poor imaging conditions such as low light and extreme weather. In comparison, thermal images that rely on the temperature of the object can reduce these external disturbances. In the past, many RGB and thermal (RGBT) vision tasks have arisen, such as RGBT tracking, RGBT person re-identification, RGBT salient object detection, etc \cite{li2018learning,zheng2021learning,9631856,tu2022rgbt,9389777}. In order to overcome the limitation brought by RGB imaging, we introduce the thermal modality to help the VOD task break through the bottleneck.
 
\begin{figure}[t]  
		\centering
		\includegraphics[width=0.5\textwidth]{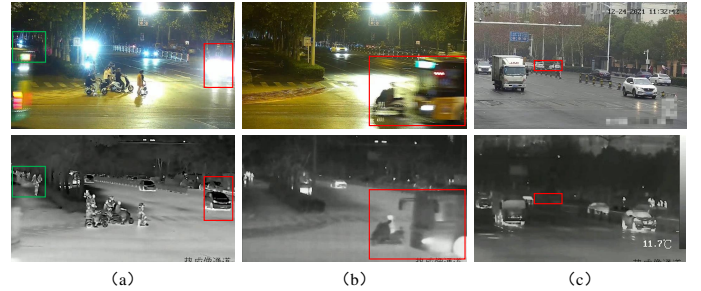}
		\caption{RGB-thermal image pairs in our dataset. Column(a) shows the advantages of the thermal image over the RGB image at night, for example, the green box gives poor illumination, and the red box shows a vehicle with strong light. Column(b) shows thermal imaging is better for the object with rapid-motion. Column(c) depicts RGB imaging's advantage for capturing details and textures of object in the daytime.} 
		\label{example}	
	\end{figure}

 Therefore, we propose a brand new task which is RGBT Video Object Detection (RGBT VOD). This new task raises two major problems. First, RGB and thermal images have their own advantages and disadvantages (as shown in Fig.\ref{example}) under different scenes and conditions. The areas marked by green boxes and red boxes in (a) show poor illumination and strong light. We can see that objects in the RGB image are already difficult to identify but the quality of thermal imaging is relatively better at this point. The marked areas in (b) show that the thermal image can better overcome the blur caused by the fast motion of the object compared to the RGB image. The areas marked by red boxes in (c) show that the RGB image contains some small objects when the light is sufficient, but these objects are lost in the thermal image because they are too far away from the camera. Thus for RGB and thermal images, how to fuse features from the two modalities with the aim of combining their strengths and avoiding their weaknesses is a key problem that we need to solve. Second, most VOD methods try their best to boost precision by using numerous auxiliary frames. However, the cost of finding auxiliary frames and maintaining a bank of auxiliary features is enormous. This problem of efficiency limits the application of VOD.

	To solve the first problem, Zhang \emph{et al.} \cite{8633180} concatenate the feature maps of RGB and thermal together and send it to the fully connected layer for prediction. Zhang \emph{et al.} \cite{ZHANG2020115756} make a deeper multi-level adaptive feature interaction between RGB and thermal images at the neck of the network. Recently, most fusion approaches \cite{8666745,9454273,9879218,8633180} still adopt multi-level interaction strategy. For example, Tu \emph{et al.} \cite{9454273} design a separate interaction module called Multi-Interactive Block (MIB). But most of the previous fusion models appear as black-box models, ignoring that some noises might arise from the multi-modal interaction. To fuse RGB and thermal modalities in a more reasonable way, we propose an Erasure-based Interaction module, which focuses on the characteristic of feature map itself. First, we use two main branches to extract the features of each modality separately. Then, inspired by Hu and Guo \emph{et al.} \cite{hu2021trash}, we design a new negative SiLU activation function to identify noise regions that do not contain objects in the thermal feature map. These noise regions are then used to erase the noise in the RGB feature map in order to extract more precise objects. To ensure fault tolerance in this process, we use the Convolutional Block Attention Module (CBAM) \cite{woo2018cbam} to guide it flexibly. In summary, our method enables full information interaction between RGB and thermal while removing the noise that may be introduced in the process.
	
	The second problem is to improve the efficiency of VOD. Sun \emph{et al.} \cite{sun2021mamba} propose a good way of updating the memory bank, which is widely used in video tasks \cite{liang2020video}, which can achieve efficient temporal feature interaction. FastVOD-Net\cite{9785382} extracts sparse keyframes to improve the speed of the method. However, the traditional approaches of relying on semantic similarity to select feature aggregation objects are error-prone if faced with scenarios where multiple objects are easily confused. In addition, these methods are not efficient enough, as they adopt complex features or depend on upstream sub-tasks. Facing the new RGBT VOD task, our method will be complex extremely if we adopt the traditional approach of maintaining memory banks for both modalities. Therefore, we propose a Temporal Proximity Enhancement (TPE) module, in which we design a local temporal window to improve efficiency further. Here we consider the strong correlation between consecutive three frames to model spatio-temporal information. Specifically, we first compute the similarity between the current frame and its adjacent two frames in terms of channel and space. The two adjacent frames are then fine-tuned by similarity weights to bring them closer to the current frame. Finally, we add up the feature maps of the two adjacent frames as the feature representation of the current frame. This approach can capture complex spatio-temporal relationships with high efficiency, especially for complex scenarios.

	After solving these two problems, we build a unified framework: Erasure-based Interaction Network (EINet) for RGBT VOD. Furthermore, we construct a RGBT VOD dataset (VT-VOD50) to evaluate the model for the new task. VT-VOD50 contains a total of 50 video sequences with varying lengths. They are captured jointly by two no-overlapping cameras in the real scene. It encompasses most of the challenges in VOD, such as lighting changes, motion blur, object occlusion, poses and shape changes, etc.
	
	We evaluate many advanced VOD methods on our VT-VOD50 and compare their performances with our EINet to demonstrate the effectiveness of our EINet, which forms the benchmark of this new RGBT VOD task.
	
	To the best of our knowledge, this is the first work to release the RGBT video object detection task and benchmark. Our main contributions can be summarized as follows:
\begin{itemize}
	\item We propose a new task called RGBT Video Object Detection for overcoming some challenges in VOD caused by RGB imaging, which can boost the effect of VOD by introducing thermal images. 
	\item We build a unified detection framework EINet for RGBT VOD. In EINet, the proposed Erasure-based Interaction module can remove the noise well from the feature maps when fusing multi-modal information.
	\item We develop the Temporal Proximity Enhancement module to improve efficiency, in which, we propose a local temporal window to model spatio-temporal information efficiently.
	\item We construct a new RGBT Video Object Detection dataset including 50 pairs of RGBT video sequences with complex background and different illumination, with manually labeled ground truth annotations. We evaluate many state-of-the-art methods on our dataset and perform extensive experimental comparisons and analyses.
\end{itemize}
	
	The rest of the paper is organized as follows: Section II presents the related work of RGBT VOD; Section III presents the structure and details of our proposed EINet; Section IV depicts our proposed RGBT VOD dataset VT-VOD50; Section V conducts detailed comparison and ablation experiments to demonstrate the effectiveness of our approach, also gives some visualization results; Section VI concludes this work and discusses the future research for this new task.
	
	\section{Related Work}
	In recent years, video object detection has been applied in many tasks such as autonomous driving, but it is limited by some poor RGB imaging circumstances such as adverse illumination conditions. In many vision tasks with RGB and thermal modalities, the information interaction and fusion between RGB and thermal modalities gradually show up its advantages for addressing many difficulties in RGB vision tasks.
	
	\subsection{Video Object Detection}
	Along with the ImageNET VID dataset \cite{russakovsky2015imagenet} being released, VOD appears and is gradually applied to numerous practical tasks\cite{taylor2016anomaly}. 
	
	DFF \cite{zhu2017deep} and FGFA \cite{zhu2017flow} are proposed to use optical flow information to aggregate more discriminative features for the current frame. However, while DFF is more concerned with speed, FGFA is more concerned with accuracy. Naturally, the performance of both models depends heavily on the effect of optical flow, which makes the algorithm still not guaranteed robust despite the sacrifice of efficiency. They do not have a good balance of speed and performance. Wang \emph{et al.} \cite{wang2018fully} improve the FGFA \cite{zhu2017flow} by combining instance-level alignment with pixel-level alignment to achieve better detection results. Wu \emph{et al.} \cite{wu2019sequence} use semantic similarity to aggregate the features from randomly sampled frames. This work \cite{wu2019sequence} focuses on global semantic information and obtains an improved performance. Nevertheless, it ignored the temporal information is still illogical. As a result, relying solely on semantic similarity to determine the object of feature aggregation is effective in simple scenarios but is still not applicable to complex multi-objective challenges on long-time video orders. The MEGA proposed by Chen \emph{et al.} \cite{chen2020memory} combines global semantic information with local localization information to help the current frame to complete the detection task. This method improves the accuracy but still requires maintaining a long range of memory bank. Gong \emph{et al.} \cite{gong2021temporal} propose Temporal ROI Align to further improve the detection by giving temporal information to the traditional ROI Align, but it still requires more support frames as an aid. In any case, the above methods exhibit limitations in effectively addressing challenges such as occlusion and similarity issues in scenarios involving multiple objects. Hence, the issue of effectively and expeditiously utilizing time-series data poses a challenge that requires resolution.
	\begin{figure*}[t]  
		\centering
		\includegraphics[width=\textwidth]{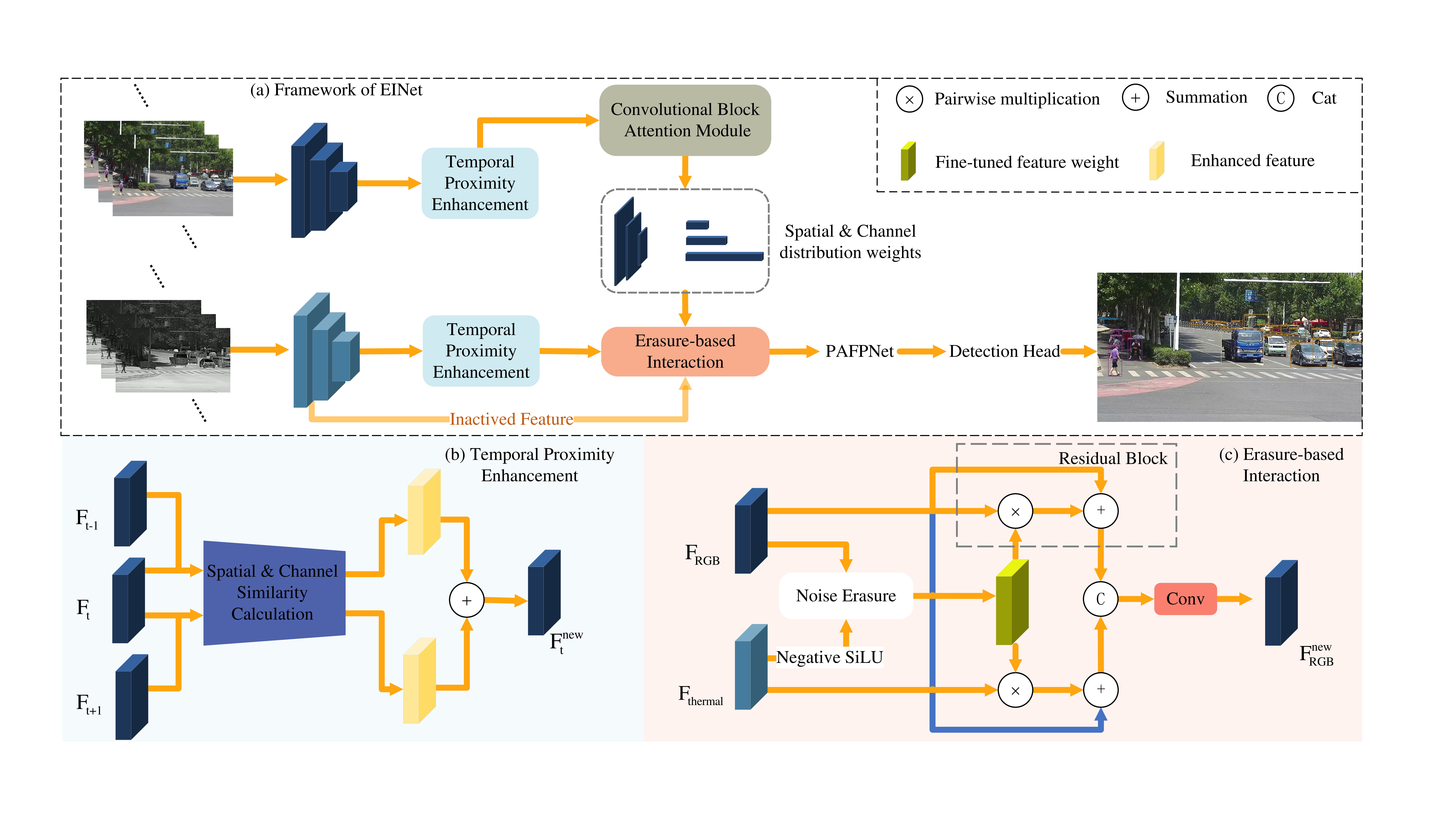}
		\caption{The network architecture of our EINet. For the inference of each frame in the video, we feed the network with a total of six photos from RGB and T. They go through two backbone branches to get their respective multi-scale features. Then we perform temporal enhancement of the current frame using the features of neighboring frames. For the critical RGBT fusion, we use the inverse activation feature of the T image for noise erasure of RGB, which is the highlight in our network. To improve the robustness of the erase operation, we also used the CBAM \cite{woo2018cbam} module to guide the process spatially and channel-wise. The fused multiscale features enter PAFPN \cite{bochkovskiy2020yolov4} for cross-scale information interaction and then three different levels of decoupled detection heads are performed to predict the results.}
		\label{network}	
	\end{figure*}
	\subsection{RGBT Feature Fusion}
	As vision research progresses and application needs evolve, the limitations of RGB images in vision tasks become more evident. These limitations include the inability to maintain image quality under harsh conditions, such as haze, rain, snow, and low light. To address these limitations, introducing another sensor modality is a reasonable choice for extending the characterization capability of the source data. Thermal images is a different spectral imaging sensors that are close to natural images. The use of RGB images in combination with thermal images is gaining attention in detection fields.
	
	Wang \emph{et al.} \cite{wang2018rgb} propose to provide different reliability weights for multi-modal data to achieve adaptive fusion. Then more methods \cite{tu2019rgb,tu2020rgbt} explore the fusion of RGB and thermal images. Tu \emph{et al.} \cite{9454273} proposed to use a global information module and multi-level RGBT interaction module to achieve a balance of semantic features and detailed information for the RGBT SOD.
	
	In addition, RGBT fusion methods have been proposed and applied to pedestrian detection task \cite{hwang2015multispectral}. The exploration of both early and late fusion on CNNs further advances the application of RGBT for pedestrian detection \cite{wagner2016multispectral}. RGBT cross-modal features have also been proposed to be combined with migration learning to achieve the goal of supervised migration \cite{xu2017learning}. That is, it satisfies the goal of supervising the training of RGB images with T images during training so that there is no need to use multispectral data during testing. This operation is used to improve the network model's ability to detect pedestrians in dim light. The illumination-aware network is introduced into the RGBT pedestrian detection task to adaptively control the input weights of the data for both modalities \cite{li2019illumination}. But the gate function here tends to lead the network to an extreme, which leads to a serious imbalance between RGB and T.
	
	In exception to the detection field, RGBT information fusion has made some progress in other vision fields such as object tracking \cite{li2018learning} and person re-identification \cite{zheng2021robust,mogelmose2013tri}.
	
	Previous efforts have been limited in terms of scenario and the objects. Recent approaches have tended to focus on resolving the challenge of changing light conditions in a scene. Furthermore, current feature fusion methods for RGB images and thermal images primarily concentrate on prominent objects such as people and cars, leading researchers to concentrate on enhancing their characteristics while disregarding the possibility that this process may enhance the surrounding noise as well.

	\section{EINet: Erasure-based Interaction Network}
	In this section, we introduce a unified detection framework called EINet for RGBT VOD, which takes advantage of the video temporal information and multi-modal imaging. We elaborate on using a negative activation function to reduce the noise of RGB feature maps presenting in the middle of the network. And EINet has a quite good effect on video object detection even with fewer frames.

	\subsection{Overview}
	We take the advanced one-stage detector YOLOX \cite{ge2021yolox} as the baseline of the proposed EINet, and utilize the classic Darknet53 \cite{redmon2018yolov3} as the backbone, which has a powerful ability for feature encoding. This is because YOLOX has a more flexible structure than YOLOV's series models such as YOLOV5 \cite{glenn_jocher_2020_4154370} and YOLOV7\cite{wang2023yolov7}. At the same time, YOLOX has a faster speed than two-stage detectors such as Faster R-CNN \cite{5736986b6e3b12023e730129} and is more suitable for RGBT VOD.
 
	The general architecture of our EINet is presented in Fig.\ref{network} (a). Features are extracted from three adjacent frames of each modality and then integrated with features from neighboring frames to enhance discriminability. RGB and thermal feature maps are then combined with the weight map generated by the attention mechanism to generate a new feature map. These multi-layer features are fed into the PAFPN network \cite{bochkovskiy2020yolov4}, which consists of a top-down Feature Pyramid Network (FPN) \cite{lin2017feature} and a bottom-up Path Aggregation Network (PAN) \cite{liu2018path}. Finally, we input the feature maps of these three layers into the detection head and perform multi-scale inference to obtain the detection result.
	
	\subsection{Multi-branch Feature Extraction}
	A multi-branch structure is employed to extract features from multiple frames of both modalities simultaneously. In order to ensure the specificity of each modality, separate parameters are employed for the feature extraction branches. In addition to the conventional method, a backbone that eliminates the activation function during the high-level feature extraction is further employed for current frame in T modality. The three original T images depicted in Fig.\ref{network} generate not only the conventional feature map, but also the inactive feature map following the backbone. The inactive feature maps are used for feature fusion between the two modalities.
	
	\subsection{Temporal Proximity Enhancement}
	In video information processing, the idea of multi-frame aggregation, which is widely used due to historical legacy problems, often requires tens of frames or even more neighboring frames to do the information aggregation. This approach is not only redundant but also does not provide a reasonable choice of the number of frames to be used. For the current mainstream VOD models that use the optical flow information and features of multiple frames before and after \cite{zhu2017flow,wang2018fully} or use them to build a long-term feature memory bank \cite{beery2020context}, we believe that this is not suitable for multiple objects in complex scenes, and the computational overhead is higher and the efficiency is low. In order to solve the above problems, we choose to use a shorter time window to handle the information fusion on the time sequence. This approach can address challenges such as object overlap and fast motion by highlighting objects with the help of feature overlap between left and right neighboring frames and the current frame. 
	
	\begin{figure}[htbp]  
		\centering
		\includegraphics[width=0.45\textwidth]{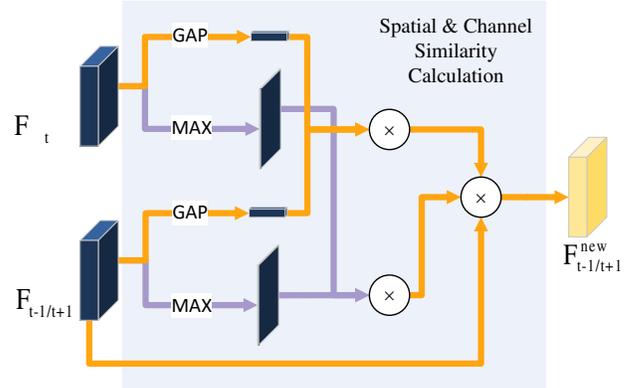}
		\caption{Structural of the Space and Channel Similarity Calculation module.}
		\label{scsc}	
	\end{figure}
	
	We take one frame before and one frame after the current frame as illustrated in Fig.\ref{network} (b). After the backbone gets the multilayer feature maps of the three frames, we get three feature maps named $F_{t-1}$, $F_t$, and $F_{t+1}$. First, we perform a global average pooling (GAP) on each of three features with size $C \times H \times W$ to obtain three tensors with size $C \times 1 \times 1$, which are $P_{t-1}$, $P_t$, and $P_{t+1}$. Similarly, we select the maximum value from the original feature map for each channel to obtain three features with size $1 \times H \times W$, which are $V_{t-1}$, $V_t$, and $V_{t+1}$, respectively. We can get the similarity tensor among feature maps by the operation of $P \times V$, and then multiply it back onto the feature map of the neighboring frame, thus we complete the similarity measure between the neighboring frame's features and the current frame's features in both channel and space. This process can be clearly seen in Fig.\ref{scsc}.
	Based on this principle, the following operations were chosen for the fusion of temporal information:
	\begin{equation}
		F^{New}_{t} = {P_{t - 1}} * {V_{t - 1}} * F_{{t - 1}} + {P_{t + 1}} * {V_{t + 1}} * F_{{t + 1}}
	\end{equation}
	where ${t-1}$ and ${t+1}$ refer to the previous and the next frame of the current frame, respectively. The $F$ refers to the feature map. By replacing the features of the current frame with those of the previous and subsequent frames in this way, we also improve the robustness of the network while ensuring performance and speed.
	\begin{figure}[htbp]  
		\centering
		\includegraphics[width=0.45\textwidth]{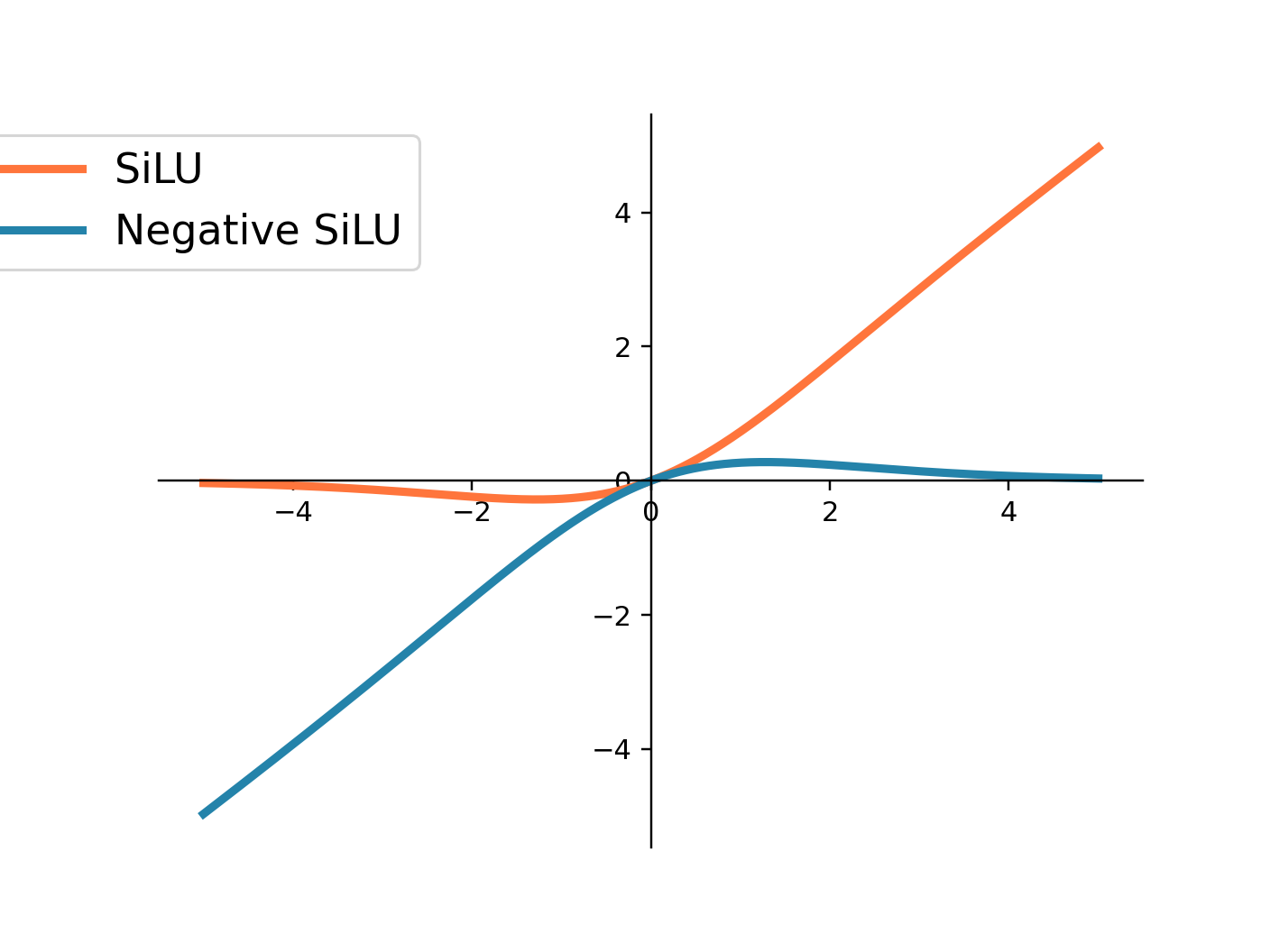}
		\caption{Diagram of SiLU and Negative SiLU functions. The orange line represents the SiLU activation function, from which we can see that its purpose is to suppress the response values in regions of the feature map that are not of interest to the network. The Negative SiLU function represented by the green line is the opposite idea, which can retain most of the negative response values of the background region without caring about the object region.}
		\label{silu}	
	\end{figure}
	
	\subsection{Erasure-based Interaction for RGBT modalities}
	When referring to RGBT data, what is often considered is how to carry out information complementation and refuse the redundant information brought by feature fusion. In our method, we expect to introduce thermal images to enrich the information of objects and remove noises in RGB images as more as possible. 
	
\begin{figure*}[b]  
		\centering
		\subfloat[]{
			\includegraphics[width=0.45\columnwidth,height=2.4cm]{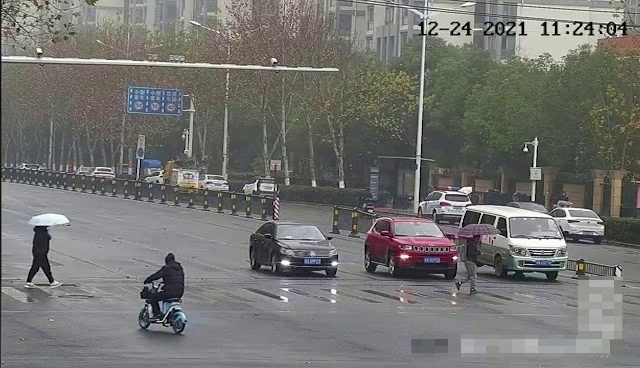}
		}
		\quad
		\subfloat[]{
			\includegraphics[width=0.45\columnwidth,height=2.4cm]{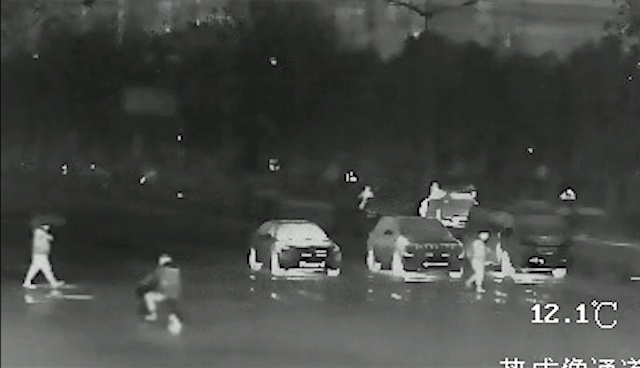}
		}
		\quad
		\subfloat[]{
			\includegraphics[width=0.45\columnwidth,height=2.4cm]{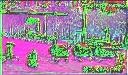}
		}
		\quad
		\subfloat[]{
			\includegraphics[width=0.45\columnwidth,height=2.4cm]{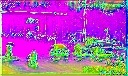}
		}
		\caption{\textbf{(a)} is an RGB image, and we can see that the foreground and the background are complexly intertwined.  \textbf{(b)} is the corresponding thermal image, whose background is relatively clean. \textbf{(c)} visualizes the output feature map of (a) after feature extraction by our backbone Darknet53. \textbf{(d)} visualizes the RGB feature map, whose noise is erased by the corresponding thermal feature map after using our negative activation function.}
		\label{Fig5}
	\end{figure*}

	The RGB image often has the complex background just as shown in Fig.\ref{Fig5} (a), so there are lots of noises around the object in RGB image's feature map like the one shown in Fig.\ref{Fig5} (c). The thermal imaging always leads to simple background as shown in Fig.\ref{Fig5} (b). Therefore, we propose to use the negative activation function to remove the noisy regions in the RGB image, together with the help of thermal image, which is also inspired by Hu \emph{et al.} \cite{hu2021trash} to some degree. The proposed approach is illustrated in Fig. \ref{network} (c).
	We apply the Negative SiLU to process the thermal feature. The SiLU function \cite{Hendrycks2016GaussianEL} and the negative SiLU function designed by us are defined as follows:
	
	\begin{equation}
		\left\{ {\begin{array}{*{20}{l}}
				{SiLU = x*\frac{1}{{1 + {e^{ - x}}}}}\\
				{SiLU_{negative} = x*\frac{1}{{1 + {e^x}}}}
		\end{array}} \right.
	\end{equation}
	Then, we obtain the feature map with negative values in the background, while the feature values in the foreground are nearly erased.
	Next, we add the background-focused feature map to the RGB feature map with noisy background to remove some noises in the background of RGB feature map. The feature map with some noises removed as illustrated in Fig.\ref{Fig5} (d) is then used to obtain a global weight after the sigmoid function. The weight is used to guide the features of both RGB and thermal modalities separately to create a residual structure that is concatenated together and sent to the PAFPN network \cite{bochkovskiy2020yolov4}. However, blind feature erasure relies heavily on the quality of the thermal image. This will make the model less robust.

	To improve the robustness of our method, we propose to use spatial attention \cite{woo2018cbam} to guide this noise erasure process. At first, we use the spatial attention module to obtain a feature map $W_{fore}$ reflecting the importance of spatial location, which mainly focuses on the foreground regions. We also utilize $W_{fore}$ to obtain another weight tensor $W_{back}$ by simply subtracting the former weight tensor from the tensor whose values are all 1. Naturally $W_{back}$ is more concerned with the noise region. Additionally, we use channel attention \cite{8578843} to focus on the crucial channels of the feature tensor. Thus we obtain a weight $W_{ch}$ through the channel attention mechanism. So the feature fusion process between the two modalities can be expressed as follows:
	\begin{equation}
		{W_{fusion}} = {\sigma _1}[{W_{fore}}*{W_{ch}}*F_{{RGB}} + 
		\nonumber
	\end{equation}
	\begin{equation}
		{W_{back}}*{W_{ch}}*{\sigma _2}(F^{Inact}_{{T}})]
	\end{equation}
	
	\begin{equation}
		F_{new} = Conv[Cat(F_{RGB} + {W_{fusion}}*F_{RGB},
		\nonumber	
	\end{equation}
	\begin{equation}
		F_{RGB} + {W_{fusion}}*F_{T})]
	\end{equation}
	where $\sigma_1$ denotes the sigmoid function, and $\sigma_2$ indicates the negative SiLU function. $F^{Inact}_{T}$ refers to the unactivated feature map of the thermal modality. To avoid the activation function filtering out the eigenvalues of the noise area, we remove some of the activation functions in the backbone, then we get the feature $F^{Inact}_{T}$.
	
	We do this fusion process by adding the foreground-enhanced RGB feature map and the foreground-excluded thermal feature map. This result is then passed through the activation function to form $W_{fusion}$ that guides the fusion. In some scenes, thermal imaging is insensitive to some small objects far away from the foreground object. So these small objects will be treated as background in the feature map. Then the $W_{fusion}$ is conducted with the biased guidance. To further improve the robustness of the fusion, we design two residual branches. Here the RGB feature maps and thermal feature maps are fine-tuned by $W_{fusion}$, then the RGB feature maps are added back to their respective branches. Finally, the feature maps produced by the two branches are concatenated. After completing the RGBT feature fusion, the dual-stream features are re-merged into single-stream features for subsequent interaction and detection. We follow the baseline approach in this part, which is not repeated here.
	
	It is worth noting that we only use the feature map of the thermal image here to help remove noise from the RGB image and do not choose a bi-directional erase operation. As mentioned earlier, this is determined by the characteristics of the two modalities of imaging. It is because of the special imaging principle of thermal images that they tend to capture the contours of the object well while leaving out the surrounding noise. At the same time, this is the disadvantage of RGB images, so all we can do is erase from thermal to RGB in one direction.

	\section{VT-VOD50 : RGBT \\Video Object Detection Datasets}
	As mentioned earlier, the existing dataset used widely for VOD algorithms is the ImageNET VID dataset \cite{russakovsky2015imagenet}. Although the ImageNET VID dataset has a large number of videos and objects, it behaves simply in terms of scenes, and has a small number of objects in each video. So this is not very good at providing a comprehensive performance review of the model. The demands of real-life applications place greater demands on data sets such as more realistic and varied scenarios, more challenges and richer picture information. Also, our proposed new task of RGBT VOD urgently requires a complete dataset to evaluate the algorithm. For these reasons we formally propose a new dataset VT-VOD50 to learn complementary information between the two modalities. The VT-VOD50 dataset consists of 50 pairs of RGBT video sequences for a total of 9449 RGBT image pairs.
	
	\begin{figure*}[b]
		\centering
		\begin{minipage}[b]{\linewidth} 
			\hfill
			\subfloat[]{
				\begin{minipage}[b]{0.18\linewidth}
					\centering
					\centering
					\includegraphics[width=\linewidth,height=1.835cm]{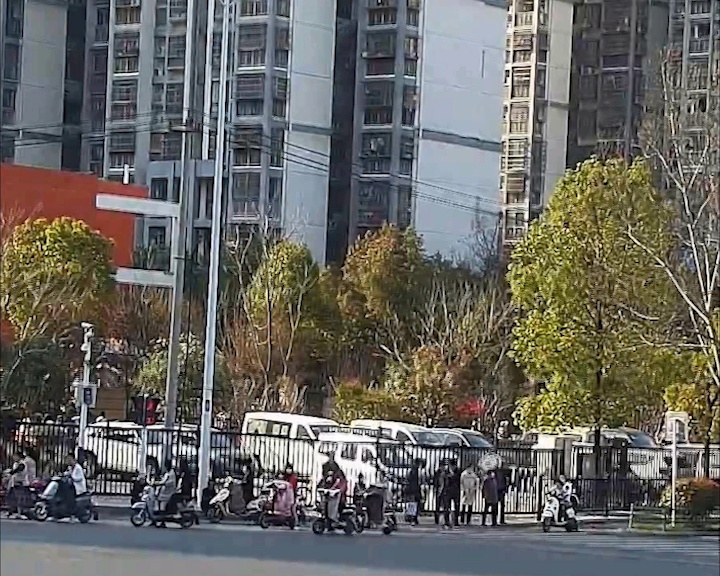}\vspace{10pt}
					\includegraphics[width=\linewidth,height=1.835cm]{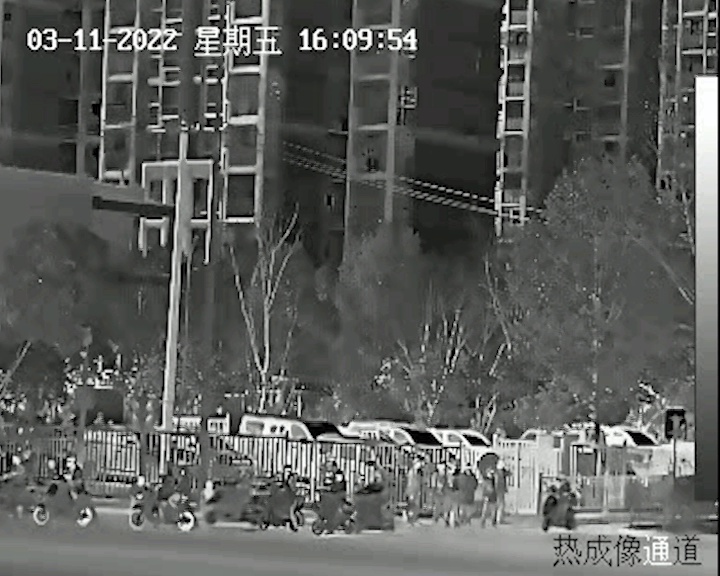}\vspace{10pt}
				\end{minipage}
			}
			\hfill
			\subfloat[]{
				\begin{minipage}[b]{0.18\linewidth}
					\centering
					\centering
					\includegraphics[width=\linewidth,height=1.835cm]{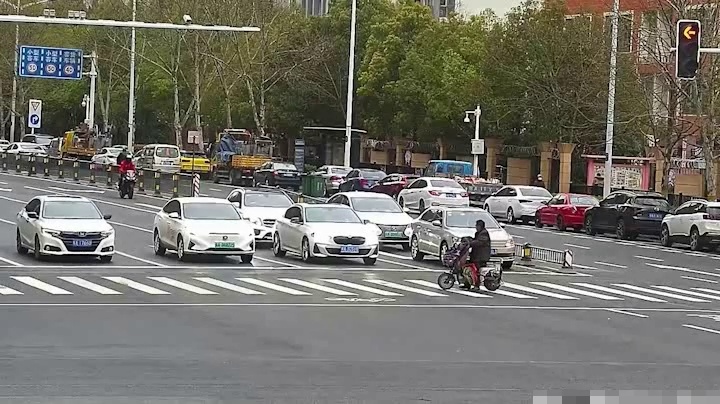}\vspace{10pt}
					\includegraphics[width=\linewidth,height=1.835cm]{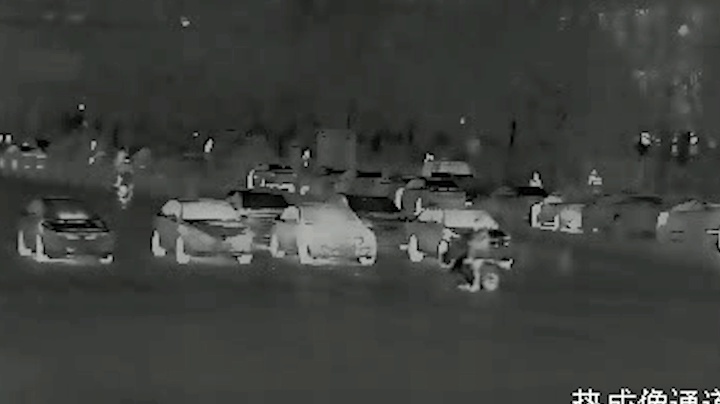}\vspace{10pt}
				\end{minipage}
			}
			\hfill
			\subfloat[]{
				\begin{minipage}[b]{0.18\linewidth}
					\centering
					\centering
					\includegraphics[width=\linewidth,height=1.835cm]{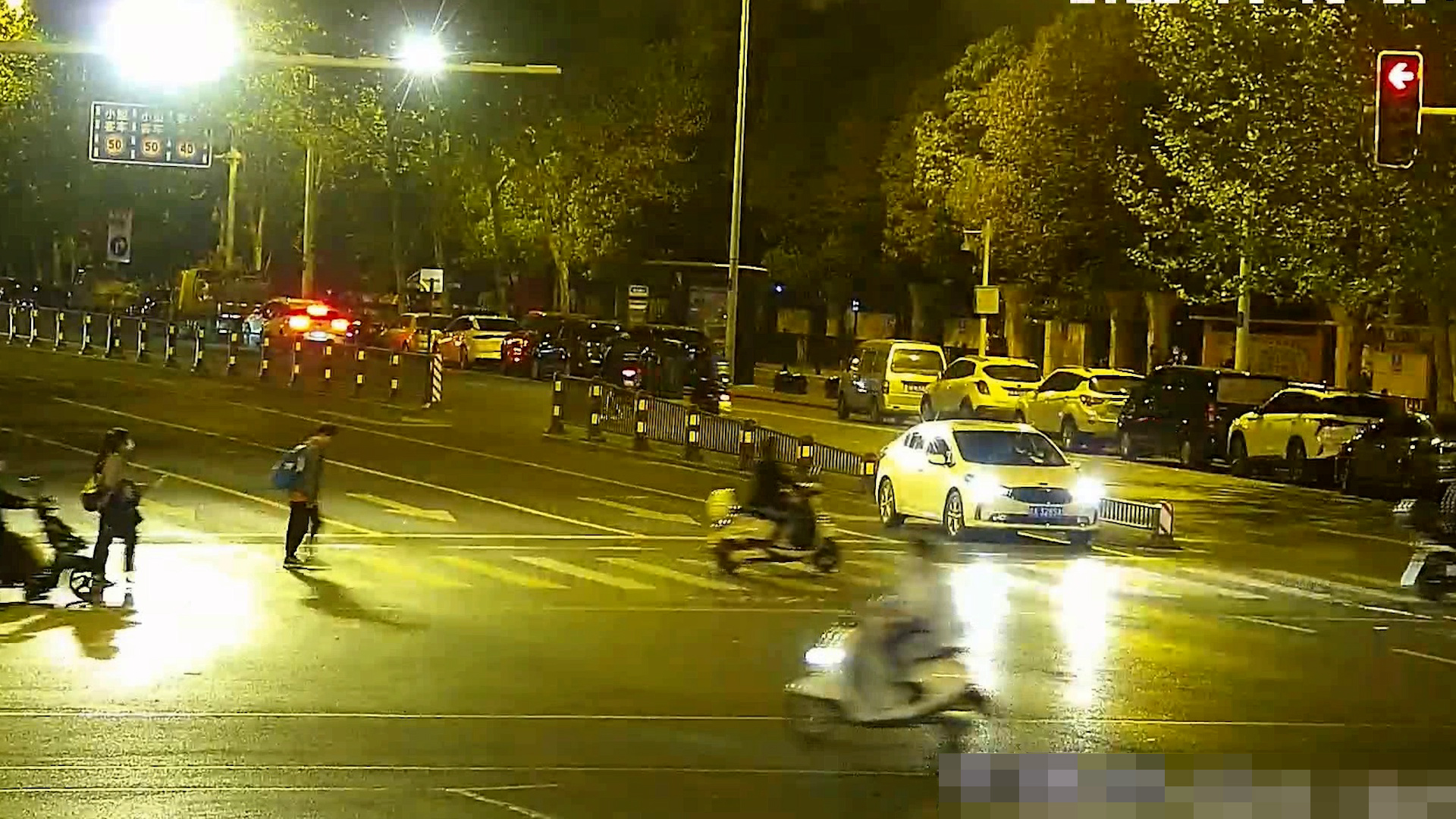}\vspace{10pt}
					\includegraphics[width=\linewidth,height=1.835cm]{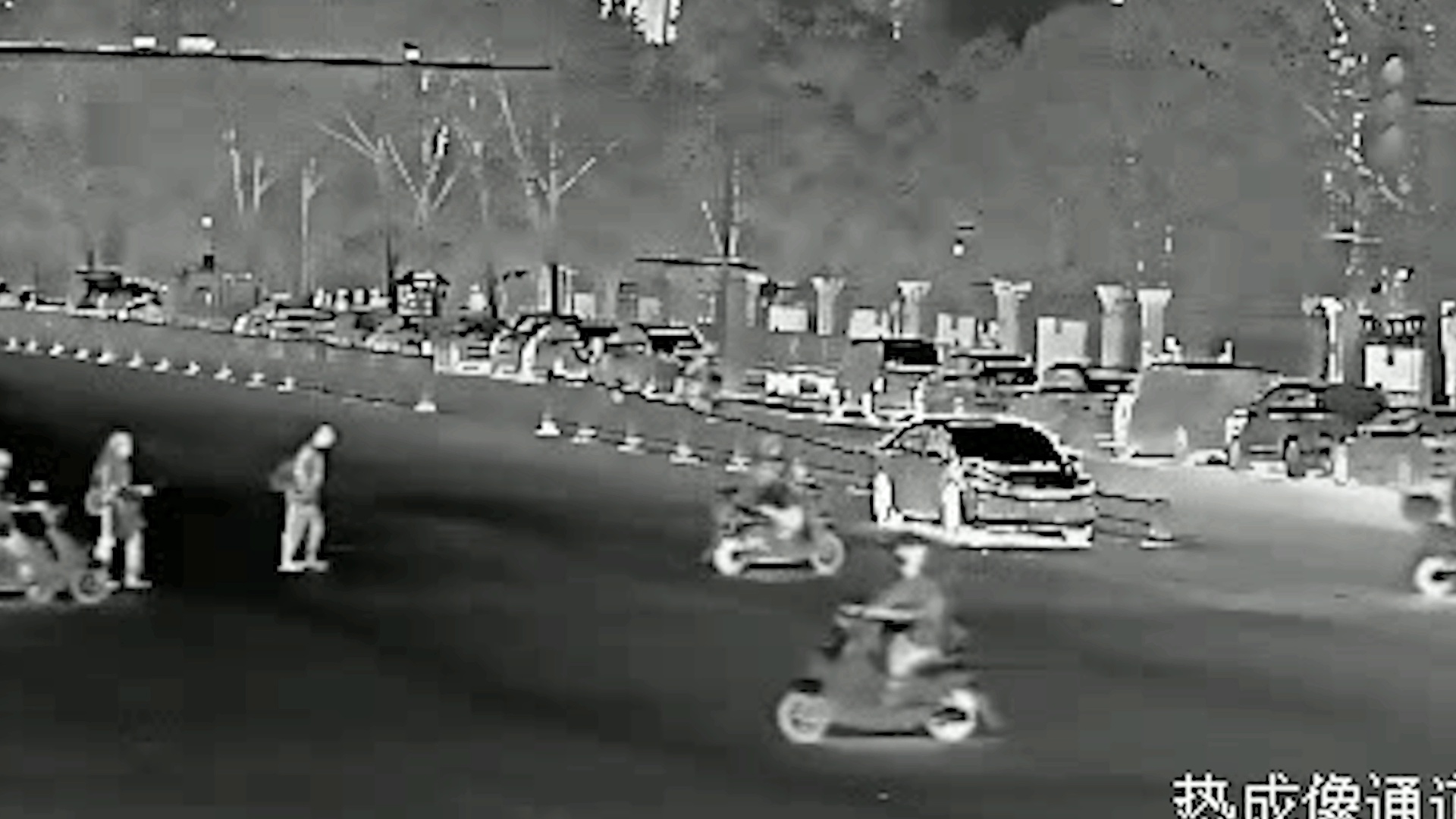}\vspace{10pt}
				\end{minipage}
			}
			\hfill
			\subfloat[]{
				\begin{minipage}[b]{0.18\linewidth}
					\centering
					\centering
					\includegraphics[width=\linewidth,height=1.835cm]{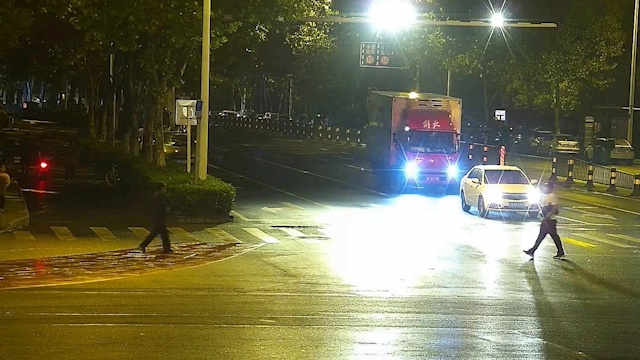}\vspace{10pt}
					\includegraphics[width=\linewidth,height=1.835cm]{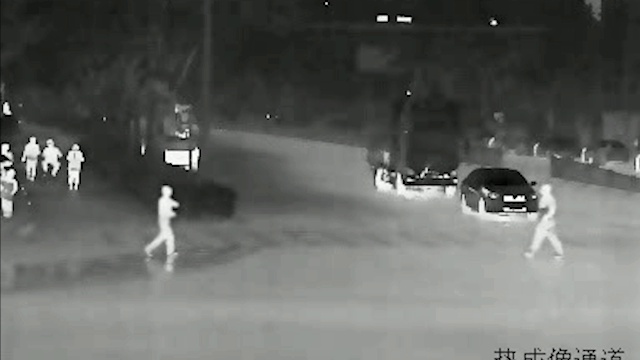}\vspace{10pt}
				\end{minipage}
			}
			\hfill
			\subfloat[]{
				\begin{minipage}[b]{0.18\linewidth}
					\centering
					\centering
					\includegraphics[width=\linewidth,height=1.835cm]{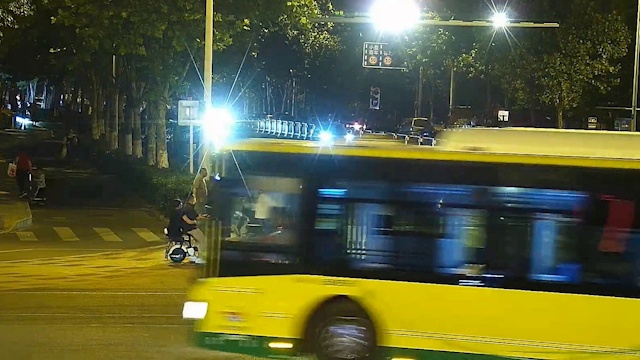}\vspace{10pt}
					\includegraphics[width=\linewidth,height=1.835cm]{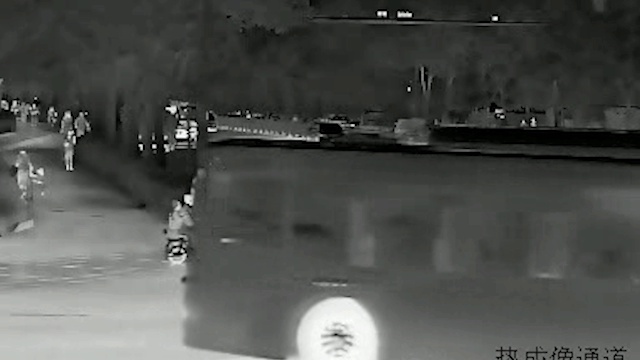}\vspace{10pt}
				\end{minipage}
			}
		\end{minipage}
		\vfill
		\caption{Some of the challenge scenarios that really exist in our dataset. \textbf{(a)} and \textbf{(b)} demonstrate the challenges of partial multi-object and severe occlusion. \textbf{(c)}, \textbf{(d)}, and \textbf{(e)} are all night scenes, but they each present different difficulties, such as \textbf{(c)} being the representative of the quick motion of the target;  targets in the \textbf{(d)} at dark light are difficult to capture by RGB; \textbf{(e)} is a motion blur.}
		\label{challenge}
	\end{figure*}

	\subsection{Data Collection}
	The videos in VT-VOD50 dataset are collected in real-life traffic scenarios. We use two cameras of Hikvision placed at the same location without overlapping to capture traffic video data of both modalities. In addition to the rich scene information as shown in Fig.\ref{example}, we collect data from cool weather to hot weather, from daytime to dusk to nighttime, to ensure that the dataset is fully oriented to real-world conditions. The dataset also has different resolutions including 640 $\times$ 368, 680 $\times$ 404, 720 $\times$ 576, 720 $\times$ 404, 1 920 $\times$ 1080, as different imaging devices have different imaging resolutions.
	
	\begin{table}[t]
        \renewcommand{\arraystretch}{1.3}
		\centering
		\caption{Distribution of the number of objects in the VT-VOD50}
        \setlength{\tabcolsep}{0.75mm}{
		\begin{tabular}{c|ccccccc|c}
			\toprule
			& Car & Van & Electromobile & Person & Bus & Truck & Bicycle & Total \\
			\midrule
			Train & 62421 & 5301  & 14625 & 45786 & 423   & 8662  & 2205  & 139423 \\
			Test & 28383 & 2292  & 5315  & 22938 & 535   & 3078  & 883   & 63424 \\
			\bottomrule
		\end{tabular}}
		\label{class}%
	\end{table}%
	\subsection{Dataset Description}
		
Our dataset includes scenes in day and at night, with the daytime scenes further subdivided into hot and cool weather categories. These scenarios present unique challenges due to their different effects on image quality between the two modalities. For instance, thermal imaging always produces high-quality images under hot weather, but low-quality images under cool weather. Furthermore, the quality of RGB images at night is less reliable or discernible compared with thermal images, due to the effects of car lights and low-light.

 In the field of VOD, the currently widely used dataset is the ImageNET VID dataset \cite{russakovsky2015imagenet}, which has lots of categories of objects and training samples. However, 
 the objects in ImageNET VID are with small number in each video sequence. In contrast, the VT-VOD50 dataset we put forward accommodates rich scene information and multiple objects across multiple categories in each video sequence.We will describe these two aspects in the following two paragraphs. 
	
	The dataset we constructed has seven common objects on the road which are car, van, electromobile, person, bus, truck and bicycle. For each category in the dataset, the distribution of the number of objects is shown in Table \ref{class}. Objects in these categories have practical challenges such as scale variation and unequal numbers. Some of the challenges present in VT-VOD50 are illustrated in Fig.\ref{challenge}, where the first row is the RGB image and the second row is the corresponding T image. In Fig.\ref{challenge}, (a) and (b) show complex situations where the object is heavily obscured and small in size; (c), (d) and (e) show the challenges in dark light scenes at night, also include the challenge of object blurring due to rapid movement as shown in (c) and (e).
	\begin{figure}[htbp]  
		\centering
		\subfloat[]{
			\includegraphics[width=\columnwidth,height=4.5cm]{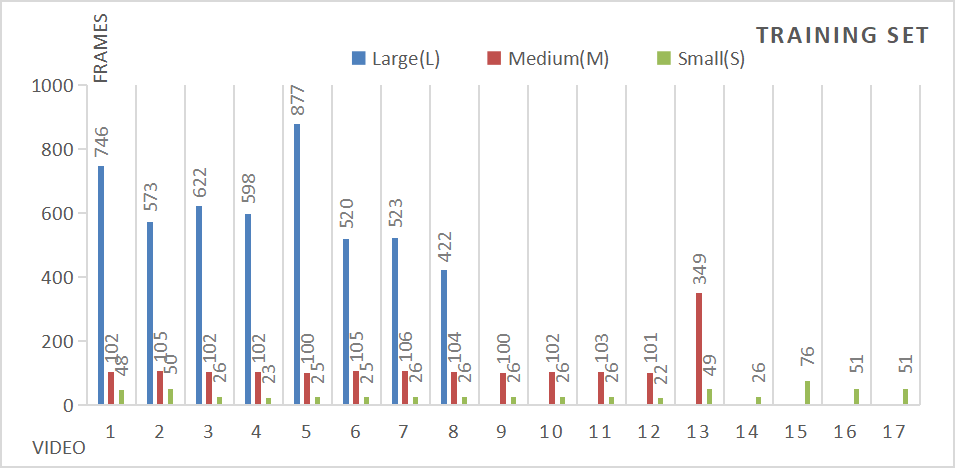}
		}
		\hfill
		\subfloat[]{
			\includegraphics[width=\columnwidth,height=4.5cm]{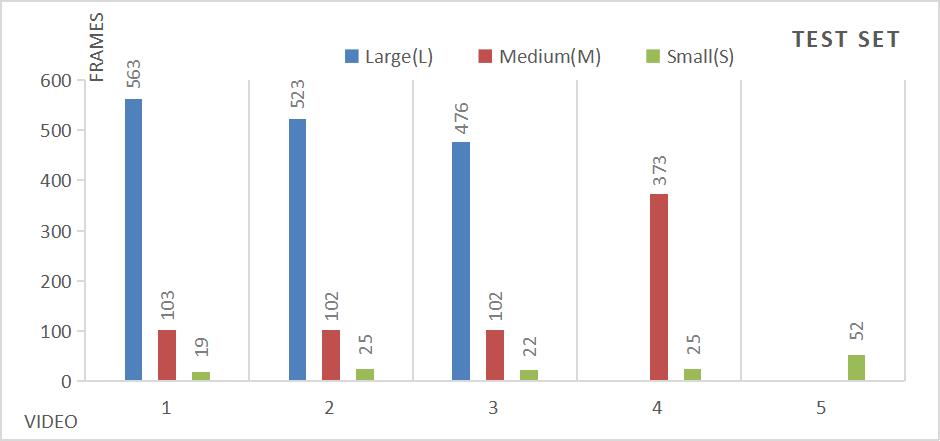}
		}
		\caption{Figure \textbf{(a)} and \textbf{(b)} show the distribution of the number and size of videos in the training and testing sets in the VT-VOD50 dataset, respectively.}
		\label{Fig3}
	\end{figure}
	We select 50 pairs of RGBT video sequences with varying length. We assign 38 pairs of RGBT videos to the training set and 12 pairs of RGBT videos to the test set as presented in Fig.\ref{Fig3}, both of which have videos with different size. In order to fully evaluate the performance of method, we also ensure that the various imaging scenarios are proportionally distributed in the training and testing sets. The organization of our dataset follows the architecture of the classical object detection dataset Pascal VOC \cite{Everingham10}, with the aim of making the data intuitive and easy to be understood. Compared with the VOD dataset ImageNET VID \cite{russakovsky2015imagenet}, which is now used by mainstream VOD algorithms of single modality for evaluation, VT-VOD50 has the following significant advantages:
	\begin{itemize}
		\item All RGB and thermal image pairs have been manually aligned to provide a good foundation for multi-modal fusion. 
		\item Our dataset reflects various challenging situations in real traffic scenes, such as poor illumination, rapid change of object scale or position, occlusion between similar objects, etc. 
		\item It provides a new platform for this new RGBT VOD task and a much more practical application scene of VOD. 
	\end{itemize}

	\section{experiments}
	In this section, we first did replacement experiments with the classical feature extractor and corroborated the experimental data by visualizing the feature maps. Then a large number of comparison experiments with classical mainstream algorithms are done, including those based on unimodal and multi-modal data. We then enumerate detailed ablation experiments to demonstrate the usefulness of each design. Finally, we show the powerful inference capability of the model with the output of the experiments.
	
	\begin{table}[htbp]
        \renewcommand{\arraystretch}{1.3}
		\centering
		\caption{We performed the replacement training using classical ResNet-50 \cite{he2016deep}. Here TI refers to the temporal information and TTI refers to the thermal images infor.}
		\begin{tabular}{c|cc|cc}
			\toprule
			Backbone & \multicolumn{2}{c|}{Darknet53} & \multicolumn{2}{c}{ResNet-50} \\
			\midrule
			& AP50(\%) & AP(\%) & AP50(\%) & AP(\%) \\
			\cmidrule{2-5}    Baseline(RGB) & 41.28 & 21.27  & 43.83 & 22.95 \\
			Baseline(T) & 27.73 & 11.93 & 29.04 & 12.45 \\
			Baseline+TI & 44.04 & 22.55 & 37.76 & 18.04 \\
			Baseline+TII(Cat) & 42.04 & 19.94 & 40.86 & 20.18 \\
			\bottomrule
		\end{tabular}%
		\label{diffback}%
	\end{table}%

	\subsection{Implementation Details}
	As the original ImageNET VID dataset \cite{russakovsky2015imagenet} can not be used to evaluate this new multi-modal task that is RGBT VOD. Therefore, we use the proposed VT-VOD50 dataset as the evaluation platform for our experiments.
	Our experiments are conducted based on the PyTorch machine-learning architecture platform. We use two NVIDIA GeForce RTX 3090s to provide arithmetic resources for the model. To ensure the fairness of the experiment, we uniformly eliminate the data enhancement operations in our method except for the horizontal flip. For input sequences with different sizes, we fill or stretch each frame to a uniform size of 640 $\times$ 640 for inputting our EINet network. In addition, we control all experiments in 150 epochs. We use Stochastic Gradient Descent (SGD) with a momentum of 0.9 and weight decay of 0.0005 to help the network learn during training. It is worth mentioning that we set the batch size to 2 for each GPU for training and inferencing.
	\begin{table*}[t]
        \renewcommand{\arraystretch}{1.3}
		\centering
		\caption{Experimental performance of EINet and other mainstream methods on VT-VOD50, best results are highlighted in bold}
		\begin{tabular}{c|c|c|c|c|c|c}
			\toprule
			Methods & Backbone & Extra Training Data & Multi-modal Information (MI)& AP50(\%)  & AP(\%) & FPS \\
			\midrule
			\multirow{4}[4]{*}{DFF (CVPR2017) \cite{zhu2017deep}} & ResNet-50 & \multirow{4}[4]{*}{} & \multirow{3}[2]{*}{} & 40.2  & 17.8  & 40.4 \\
			& ResNet-101 &       &       & 39.5  & 17.6  & 40.9 \\
			& ResNet-X101 &       &       & 34.4  & 13.7  & 36.5 \\
			\cmidrule{4-4}          & ResNet-50 &       & \checkmark     & 33.5  & 14.1  & 43.3 \\
			\midrule
			\multirow{4}[4]{*}{FGFA (ICCV2017) \cite{zhu2017flow}} & ResNet-50 & \multirow{4}[4]{*}{} & \multirow{3}[2]{*}{} & 40.5  & 17.6  & 9 \\
			& ResNet-101 &       &       & 43.6  & 20.1  & 7.5 \\
			& ResNet-X101 &       &       & 41    & 17.9  & 7 \\
			\cmidrule{4-4}          & ResNet-50 &       & \checkmark     & 35.1  & 15.8  & 9.2 \\
			\midrule
			\multirow{4}[3]{*}{SELSA (ICCV2019) \cite{wu2019sequence}} & ResNet-50 & \multirow{4}[3]{*}{} & \multirow{3}[2]{*}{} & 43.5  & 21.2  & 10.5 \\
			& ResNet-101 &       &       & 43.9  & 21.2  & 9.8 \\
			& ResNet-X101 &       &       & 43.1  & 19.6  & 7.8 \\
			\cmidrule{4-4}          & ResNet-50 &       & \checkmark     & 39.4  & 17.4  & 10.6 \\
			\midrule
			\multirow{4}[2]{*}{Temporal ROI Align (AAAI2021) \cite{gong2021temporal}} & ResNet-50 & \multirow{4}[2]{*}{} & \multirow{3}[1]{*}{} & 41.8  & 19.9  & 5.1 \\
			& ResNet-101 &       &       & 43    & 20.8  & 5 \\
			& ResNet-X101 &       &       & 40.4  & 18.3  & 4.4 \\
			\cmidrule{4-4}          & ResNet-50 &       & \checkmark     & 38    & 17    & 5.2 \\
			\midrule
			\multirow{2}[1]{*}{TransVOD (ACM MM2021) \cite{he2021end}} & ResNet-50 & \multirow{2}[1]{*}{COCO \cite{lin2014microsoft}} & \multirow{2}[1]{*}{} & 40.9  & 21.5  & 28.9 \\
			& ResNet-101 &       &       & 36.7     & 20.4     & 23.5 \\
                \midrule
                \multirow{2}[1]{*}{TransVOD++ (TPAMI2022) \cite{zhou2022transvod}} & \multirow{2}[1]{*}{Swin-B} & \multirow{2}[1]{*}{COCO \cite{lin2014microsoft}} & \multirow{2}[1]{*}{} & 46.0  & 25.0  & 8.5 \\
                \cmidrule{4-2}
			& & &    \checkmark   & 44.4     & 23.7     & 8.5 \\
			\midrule
			EINet (w/o MI) & Darknet53 &       &      & 44.04 & 22.55 & 204.2 \\
			\midrule
			EINet & Darknet53 &       & \checkmark     & \textbf{46.32} & \textbf{23.96} & \textbf{92.59} \\
			\bottomrule
		\end{tabular}%
		\label{compara}%
	\end{table*}%
	
	\begin{figure}[htbp]  
		\centering
		\includegraphics[width=0.48\textwidth]{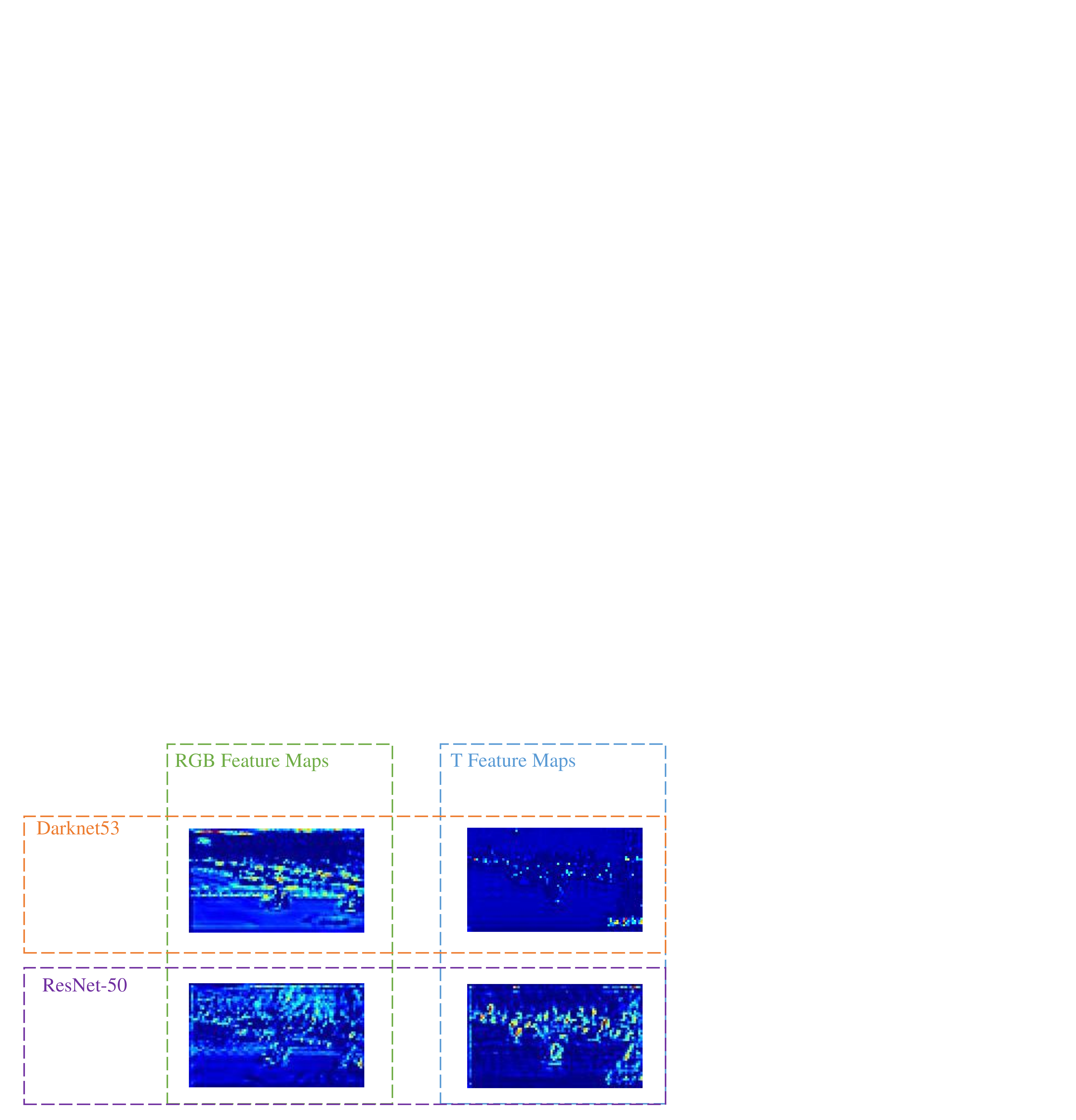}
		\caption{Visualization results of feature extraction by Darknet53 and ResNet-50 \cite{he2016deep} for RGB and Thermal images. We can see that the performances of the two different feature extractors are not consistent. Specifically, Darknet53 is better for extracting object features in the RGB image. ResNet-50 not only lacks detailed information such as contour and texture in RGB images but also brings more noises to feature extraction for thermal images.}
		\label{backbone}	
	\end{figure}
	
	\subsection{Impact of different backbones}
	Without limiting to the intrinsic architecture of YOLOX \cite{ge2021yolox}, we also use the classical backbone network ResNet-50 \cite{he2016deep} for feature extraction as a instead, the corresponding experimental results are shown in Table \ref{diffback}. The features extracted from the two modalities by two different backbone networks are visualized and presented in Fig.\ref{backbone}.
	
	First, we can see that the performances of Darknet53 and ResNet-50 are similar to each other for RGB images, and the latter performs slightly better than the former for thermal images, which can also be seen from the two feature maps of the thermal image in Fig.\ref{backbone}, where the feature map from ResNet-50 contains more discriminative information. Next, when we introduce more temporal information on RGB images, the features on Darknet53 are improved. However, the features obtained by ResNet-50 perform not so satisfied. We can see from the feature maps of RGB in Fig.\ref{backbone} that the feature maps obtained by ResNet-50 include more complex noises in the background. Finally, we get comparable performance when we concatenate and fuse the two modalities after extracting features with different feature extractors.
	
	Overall, Darknet53 performs better for RGB images, as evidenced by a more apparent distinction between foreground and background, also by a high enough focus on the object. However, ResNet-50 performs better for thermal images, capturing even distant objects. Considering the performance in all cases, we choose Darknet53 as the feature extractor of EINet without replacement.
	
	\subsection{Comparative Experiments}
	To demonstrate the effectiveness of EINet, we compare it with the classical and popular VOD methods DFF \cite{zhu2017deep}, FGFA \cite{zhu2017flow}, SELSA \cite{wu2019sequence}, Temporal ROI Align \cite{gong2021temporal}, TransVOD \cite{he2021end} and TransVOD++ (single-frame) \cite{zhou2022transvod}. As Table \ref{compara} (where AP refers to the average mAP that represents the detector over different IoU thresholds (from 0.5 to 0.95 in steps of 0.05)), We compare performances of the different methods on RGB data and RGBT data. It should be noted that we only use the data in VT-VOD50 for training other methods, except for TransVOD and TransVOD++ which use the COCO dataset \cite{lin2014microsoft} as extra training data. This is because the Transformer-based approach, while capable of learning, requires vast amounts of data to support it.
	
	From Table \ref{compara}, it seems that our EINet has overall excellent performance and better efficiency compared with the current state-of-the-art methods. EINet also improved its best performance relative to the next best TransVOD++ \cite{zhou2022transvod} by 0.32\% on the AP50 metric. The detection speed of EINet is almost eleven times faster than that of TransVOD++, which is reflected in the FPS metric. 92.59 FPS achieved by EINet can meet the needs of almost every scenario in real life.
	
	From a quantitative perspective again, our temporal proximity enhancement fusion method also outperforms the suboptimal SELSA \cite{wu2019sequence} by 0.14\% on the AP50 metric when trained using only RGB images. At the same time, our detection speed is improved by a factor of nearly 20. Based on the above, we provide a method for introducing thermal image data for the four comparison methods except TransVOD \cite{he2021end}, i.e., fusing RGB and thermal information at the entrance of the network by summing the source data of the two modalities in pixel-wise correspondence. We see through Table \ref{compara} that the performances of all five comparison methods after introducing multi-modal information are degraded, because of the heterogeneity between the modalities or a certain contradiction to the feature learning of the network, which illustrates the effectiveness of our designed multi-modal fusion method and the necessity of introducing thermal modality. TransVOD is not shown here, because its performance is too poor.
	Overall, EINet accomplishes a more accurate detection with faster speed.
	
	In addition, Table \ref{compara} shows some interesting results, such as SELSA's ability to outperform Temporal ROI Align \cite{gong2021temporal} on VT-VOD50 instead. We believe that the latter is designed for sparse objects like in the ImageNet VID dataset \cite{russakovsky2015imagenet}, and when faced with complex scenarios with many objects in real life, the connections between redundant suggestion frames may be mismatched, bringing lower performance instead. For the slightly earlier methods such as DFF \cite{zhu2017deep}, the original intention of limiting the feature modeling capability of the network in pursuit of detection speed also led to increasingly poor inference accuracy when increasing the backbone network; FGFA \cite{zhu2017flow} in the same period chose another approach that pursued performance but gave up efficiency, and the overall performance on the VT-VOD50 dataset also matched the original intention of both methods.
	
	\subsection{Ablation Studies}
	In order to verify the effectiveness of each module in our proposed method, we conduct several ablation experiments separately as shown in Table \ref{ablation}.
        \begin{table}[htbp]
  \renewcommand{\arraystretch}{1.3}
  \centering
  \caption{Ablation Studies for Temporal Proximity Enhancement (TPE) and Multi-modal Information (MI)}
    \begin{tabular}{l|ccrc}
    \toprule
    Models & Params(M) & Gflops & \multicolumn{1}{l}{AP50(\%)} & FPS \\
    \midrule
    Baseline & 8.94     & 41.83     & 41.28 & 346.0 \\
    +TPE   & 8.94     & 48.83     & 44.04 & 204.2 \\
    +MI(cat) & 19.35     & 49.12     & 42.04 & 188.6 \\
    +MI(Erasured Interaction) & 23.65     & 60.18     & 43.97 & 142.2 \\
    EINet & 23.65     & 104.31    & 46.32 & 92.6 \\
    \bottomrule
    \end{tabular}
  \label{ablation}
\end{table}%
	
	\begin{figure*}[t]  
		\centering
		\includegraphics[width=\textwidth]{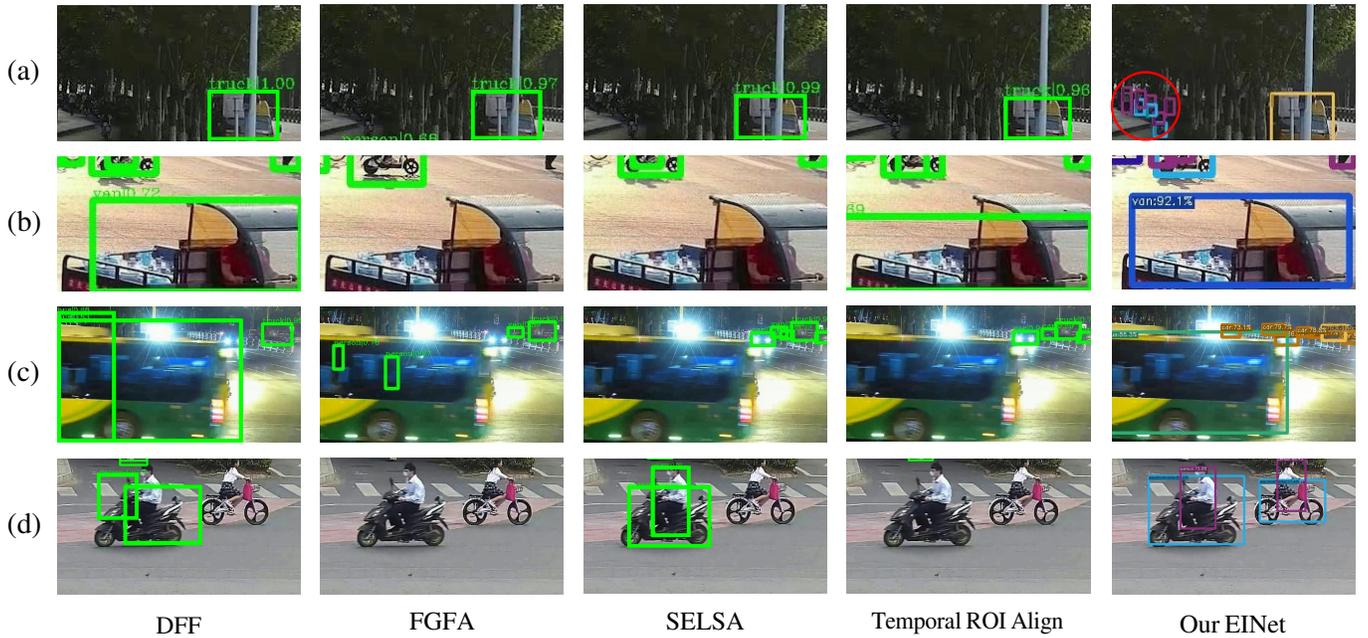}
		\caption{Visualization images of inference results. We test four methods DFF \cite{zhu2017deep}, FGFA \cite{zhu2017flow}, SELSA \cite{wu2019sequence} and Temporal Roi Align \cite{gong2021temporal} together with EINet under several different conditions, including multi-object day and night scene, night scenario with multiple challenges and high-resolution video. In these cases, EINet has a significant advantage over other models in terms of the number of detected objects and the accuracy of the detected locations. At the same time, EINet does not have the problem of redundancy of detection anchors of other methods.}
		\label{visual}	
	\end{figure*}

	We will explain each part of the experiment line by line as follows. 1): First is our baseline algorithm on VT-VOD50.
	On RGB data, our baseline shows an accuracy of 41.28\% for detection at IOU=0.5, also as a single-frame detector. 2): Then we add Temporal Proximity Enhancement (TPE) module to the RGB data, i.e., we use a total of three frames to train for each current frame, achieving an improvement of 2.76\% compared with the single-frame detection method. 3): We introduce thermal images as the second modality without using the temporal information, for which we do two groups of fine-grained experiments. In the first group, the RGB and thermal features extracted from the two backbone networks are concatenated together and convolved to fuse the two modalities, but this does not seem to be great enough from the experimental results (42.04\%). The second group is trained using the full RGBT fusion structure guidance, and the results meet our expectations achieving a performance that exceeds the unimodal data by 2.69\%, while the inference speed is also reduced to 142.15 FPS due to the introduction of thermal data. 4): Finally we jointly use the temporal information and thermal images, the performance of our EINet is further improved reaching 46.32\% on the AP50 evaluation index and the inference speed is around 93 frames per second on average. Experiments have proven that each module we designed in our EINet plays its role.

 \begin{table}[htbp]
  \renewcommand{\arraystretch}{1.3}
  \centering
  \caption{Ablation Studies for different choices of auxiliary frames. ${F_{t}}$ denotes the current frame, ${F_{t-1}}$ and ${F_{t-2}}$ denote the adjacent first two frames, and ${F_{t+1}}$ and ${F_{t+2}}$ denote the adjacent next two frames in that order.}
  \setlength{\tabcolsep}{1.2mm}{
    \begin{tabular}{c|c|c|c|cc|c}
    \toprule
    Groups & Frames used & \makecell[c]{Params\\(M)} & Gflops & AP50(\%)  & AP(\%) & FPS \\
    \midrule
    a     & ${F_{t-2}}$,${F_{t}}$ & \multirow{4}[8]{*}{8.94} & \multirow{4}[8]{*}{37.8} & 43.41 & 22.28 & \multirow{4}[8]{*}{267.4} \\
\cmidrule{1-2}\cmidrule{5-6}    b     & ${F_{t-1}}$,${F_{t}}$ &       &       & 43.45 & 20.98 &  \\
\cmidrule{1-2}\cmidrule{5-6}    c     & ${F_{t+1}}$,${F_{t}}$ &       &       & 43.78 & 22.62 &  \\
\cmidrule{1-2}\cmidrule{5-6}    d     & ${F_{t+2}}$,${F_{t}}$ &       &       & 43.36 & 22.19 &  \\
    \midrule
    e     & ${F_{t-2}}$,${F_{t-1}}$,${F_{t}}$ & \multirow{4}[8]{*}{8.94} & \multirow{3}[6]{*}{48.83} & 40.69 & 20.57 & \multirow{3}[6]{*}{204.2} \\
\cmidrule{1-2}\cmidrule{5-6}    f     & ${F_{t-1}}$,${F_{t+1}}$,${F_{t}}$ &       &       & 44.04 & 22.55 &  \\
\cmidrule{1-2}\cmidrule{5-6}    g     & ${F_{t+1}}$,${F_{t+2}}$,${F_{t}}$ &       &       & 40.67 & 20.12 &  \\
\cmidrule{1-2}\cmidrule{4-7}    h     & \makecell[c]{${F_{t-2}}$,${F_{t-1}}$\\${F_{t}}$,${F_{t+1}}$,${F_{t+2}}$}&       & 70.9  & 44.65 & 22.60  & 112 \\
    \bottomrule
    \end{tabular}%
  \label{ablation-time}}
\end{table}%

	In addition, we do ablation experiments for the selection of local temporal windows, as shown in Table \ref{ablation-time}. From the experimental results, we can analyze that the accuracy of groups a, b, c, and d is slightly lower because only the information of a single auxiliary frame is used and the feature aggregation effect is limited. Groups e and g are also not as optimal because they use ${F_{t-2}}$ and ${F_{t+2}}$ frames that are slightly further away from the ${F_{t}}$. The best result is achieved by group h. But by this time the speed of the network has dropped considerably. So we chose the f-group approach, which balances speed with performance. We believe that this is the result of sufficient aggregation of spatio-temporal information and therefore we adopt this approach in EINet.

	\subsection{Visualization}
	As presented in Fig.\ref{visual}, we use the comparison methods and EINet to do a series of visual inferences and select some representative samples. For the sake of simplicity, the images shown are cropped out of the original video frames.
	
	The images in (a) of Fig.\ref{visual} show us inference for a common daytime situation. We see that the first four comparison methods fail to detect the pedestrian in the shadows. But these are well captured by EINet with the help of our design that introduces thermal modality and augmentation of features by pulling in close in the temporal. As shown in the area we have highlighted it with a red circle.
	
	Looking again at the images in group (b), it is a daytime scene. There is a three-wheeled van about to leave the frame in the lower part. Both FGFA \cite{zhu2017flow} and SELSA \cite{wu2019sequence} missed the van due to the incompleteness of the object. The Temporal Roi Align \cite{gong2021temporal} also detects the presence of the van, but gives a bounding box with a large offset, which we believe is that the model's modeling of temporal information is not robust enough. DFF \cite{zhu2017deep} and EINet are well aware of the presence of the van. Even with this serious challenge, EINet gives a more confident score of 92.1\% than DFF.
	
	Following images of group (c) again, we pick a scene at night. Challenges included in the scene include overexposed areas and dark areas. More specifically it contains challenges such as fast motion and motion blur. Due to the excessive pursuit of detection speed, We can see that the post-processing of DFF \cite{zhu2017deep} is not particularly well done, so it gives redundant results such as two buses, and the detection locations given are not credible. FGFA \cite{zhu2017flow} mistakenly takes the shadow section inside the bus as two people during the inference. Unfortunately, neither SELSA \cite{wu2019sequence} nor Temporal Roi Align \cite{gong2021temporal} detect the presence of such an obvious bus. Looking at our EINet again still gives a near-perfect bounding box for this fast-moving bus. On the other hand, for the cars parked on the street in the upper right corner of the image, EINet's detection is also significantly better than other methods.
	
	The resolutions of images in (a), (b), and (c) stay within 1000*1000, and the quality of the images is getting higher and higher with the development of realistic technology and equipment. The images in (d) are the video we pick out with a resolution of 1920*1080, which is in line with the current application needs and trends. We can see from Fig.\ref{visual} that the FGFA \cite{zhu2017deep} and Temporal Roi Align \cite{gong2021temporal} do not make effective detection at such a resolution. And although DFF \cite{zhu2017deep} and SELSA \cite{wu2019sequence} give the detection results, they are clearly not credible. But for such a scenario, EINet still works normally and is not affected.
	
	To summarize, the compared unimodal algorithms are still not robust enough in the face of challenges, which is a limitation imposed by a single data source. The difference is that our RGBT data can break this limitation better and lead to more robust recognition information in extreme challenge scenarios.

    \subsection{Future Work}
    We take the RGBT VOD as our long-term task, because the thermal modality can be used to assist RGB modality to boost VOD effectively. So we are preparing a more comprehensive and larger dataset for RGBT VOD, which is planed to have more than 500 pairs of RGBT videos. We also plan to label each video pair with different challenging attributes such as dark lighting, motion blur, and bad weather, etc.
    In addition, we will work on more efficient and effective RGBT VOD models, that can boost RGBT VOD being applied to more multi-modal video tasks.

    \section{Conclusion}
    In this paper, we introduce a novel task called RGBT Video Object Detection, which expands upon the existing field of Video Object Detection (VOD) and enhances its practicality. We present a new detection architecture called EINet, specifically tailored for this task. The EINet architecture leverages the temporal information embedded in the video and complementary multi-modal information, while effectively managing computational costs to achieve optimal performance and speed that meets practical application requirements.
	
	Additionally, we introduce a novel dataset, VT-VOD50, specifically designed for the RGBT VOD task. The VT-VOD50 dataset collects data from real-life scenarios under various lighting and environmental conditions, to further study the complementary strengths of two modalities in different scenarios. Comparative analysis between EINet and existing mainstream detection algorithms are carried out on the VT-VOD50 dataset, which reveals that EINet offers superior performance. Lastly, we envisage that this research will stimulate further studies in both application-driven and online detection domains.
	
	\bibliographystyle{IEEEtran}
	\bibliography{IEEEabrv,mylib}
\end{document}